\documentclass{article}

\usepackage[final]{neurips_2024}

\usepackage{authblk}
\usepackage{natbib}
\usepackage[utf8]{inputenc} %
\usepackage[T1]{fontenc}    %
\setcitestyle{square, comma, numbers,sort&compress, super}
\usepackage{microtype}
\usepackage{hyperref}
\usepackage{url}
\usepackage{booktabs}
\usepackage{graphicx}
\usepackage{amsmath,amsthm,amsfonts,amssymb,bm}
\usepackage{enumitem}
\usepackage{float}
\usepackage{multirow}
\usepackage{times}
\usepackage{latexsym}
\usepackage[linesnumbered, ruled]{algorithm2e}
\usepackage{dsfont}
\usepackage{mdframed}
\usepackage{subcaption}
\usepackage[normalem]{ulem}
\usepackage[dvipsnames]{xcolor}
\usepackage{soul}
\usepackage{framed}
\usepackage{varioref}
\usepackage{float}
\usepackage{dashrule}
\usepackage{pifont}
\usepackage{helvet}
\usepackage{placeins}

\definecolor{green}{rgb}{0.1,0.1,0.1}
\definecolor{chocolate}{HTML}{D2691E}
\definecolor{maroon}{HTML}{DC267E}
\definecolor{indigo}{HTML}{4B0082}
\definecolor{green}{HTML}{008000}
\definecolor{lightblue}{HTML}{6C8EBF}
\definecolor{lightgreen}{HTML}{0E8088}
\definecolor{orange}{HTML}{FE6000}
\definecolor{cadmiumgreen}{rgb}{0.0, 0.42, 0.24}

\usepackage{amssymb}%
\usepackage{pifont}%

\makeatletter
\newcommand*\myfontsize{%
  \@setfontsize\myfontsize{8}{9}%
}
\makeatother

\makeatletter
\newcommand*\mysmallfontsize{%
  \@setfontsize\mysmallfontsize{7.4}{8.4}%
}
\makeatother

\usepackage{booktabs}

\newcommand{\ours}{\textsc{IQA-Eval}}

\usepackage{adjustbox}

\newcommand{\myskip}[1]{}

\newcommand{\barry}{\textbf{Barry Wang}\thanks{Work done while at the Department of Computer Science, Cornell University.}\space\space}

\newcommand{\modelname}[1]{{\fontfamily{lmtt}\selectfont{#1}}}
\newcommand\gpttextdavinci{\modelname{Text\-Davinci}\xspace}
\newcommand\gpttextbabbage{\modelname{Text\-Babbage}\xspace}
\newcommand\gptdavinci{\modelname{Davinci}\xspace}

\newcommand\llama{\modelname{Llama2}\xspace}
\newcommand\zephyr{\modelname{Zephyr}\xspace}
\newcommand\gptthree{\modelname{GPT3.5}\xspace}
\newcommand\chatgpt{\modelname{ChatGPT}\xspace}
\newcommand\gptfour{\modelname{GPT4}\xspace}
\newcommand\claude{\modelname{Claude}\xspace}

\usepackage{wrapfig}

\title{\ours: Automatic \underline{Eval}uation of Human-Model \underline{I}nteractive \underline{Q}uestion \underline{A}nswering}

\author[1]{\textbf{Ruosen Li}}
\author[1]{\textbf{Ruochen Li}}
\author[2]{\barry}
\author[1]{\textbf{Xinya Du}}

\affil[1]{Department of Computer Science, University of Texas at Dallas}
\affil[2]{Department of Computer Science, Carnegie Mellon University}

\affil[1]{\texttt{\{ruosen.li, ruochen.li, xinya.du\}@utdallas.edu}}
\affil[2]{\texttt{barryw@cs.cmu.edu}}

\begin{document}

\maketitle

\begin{abstract}
To evaluate Large Language Models (LLMs) for question answering (QA), traditional methods typically focus on assessing single-turn responses to given questions.
However, this approach doesn't capture the dynamic nature of human-AI interactions, where humans actively seek information through conversation\footnote{Details are in Appendix \ref{sec:accurate}}.
Recent works in human-computer interaction (HCI) have employed human evaluators to conduct interactions and evaluations, but they are often prohibitively expensive and time-consuming to scale. 
We introduce an automatic evaluation framework \ours\ to achieve \underline{I}nteractive \underline{Q}uestion \underline{A}nswering \underline{Eval}uations\footnote{\url{https://github.com/du-nlp-lab/IQA-Eval}}, more specifically, we introduce a LLM-based Evaluation Agent (LEA) that can:
(1) simulate human behaviors to generate interactions with IQA models;
(2) automatically evaluate the generated interactions.
Moreover, we propose assigning personas to LEAs to better simulate groups of real human evaluators.
We show that: (1) our evaluation framework with GPT-4 (or Claude) as the backbone model achieves a high correlation with human evaluations on the IQA task; (2) assigning personas to LEA to better represent the crowd further significantly improves correlations.
Finally, we use our automatic metric to evaluate five recent representative LLMs with over 1000 questions from complex and ambiguous question answering tasks, which comes with a substantial cost of \$5k if evaluated by humans.

\end{abstract}

\section{Introduction}

The advent of Large Language Models (LLMs) has significantly advanced the field of natural language processing (NLP), enabling systems to perform a wide range of tasks with remarkable proficiency \citep{zhao2023survey, wei2022emergent, yang2024logicalreasoningnaturallanguage, du2024making, jing2024dsbenchfardatascience}.
Among these tasks, question answering (QA) has emerged as a critical and representative goal-oriented application, demonstrating the potential of LLMs to generate informative responses as an assistant~\citep{biancofiore2024interactive}.
Multiple methods have been proposed to enhance the faithfulness and explainability of generated information~\citep{wei2022chain, long2023large, li2023leveraging, Du_2024}. Beyond developing these methods, rigorous evaluation of the generated outputs is also crucial.

\begin{figure}[h]
    \centering
    \includegraphics[width=1\columnwidth, scale=1.5]{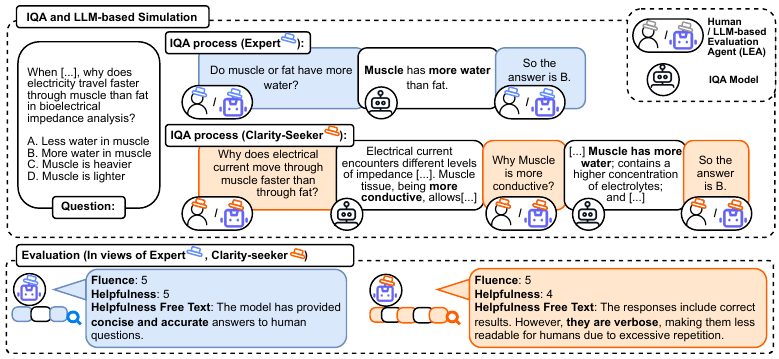}
    \caption{An example of human-model interactive question answering (IQA) and our automatic evaluation (\ours).
    The two interactions occur with two types of personas in humans (or LLM-based evaluation agents): \textbf{\textcolor{lightblue}{Expert}} and \textbf{\textcolor{orange}{Clarity-seeker}}, and are evaluated by humans or agents with corresponding personas.
    The IQA model only responds to the immediately preceding prompt without further contexts like the question itself (leftmost in the Figure).}
    \label{fig:sample}
    \vspace{-5mm}
\end{figure}

Accurate and consistent evaluation helps researchers understand existing LLM's capacities and emerging human-LLM QA interactions \citep{chang2023survey,lin2023llm, chang2023survey}.
Traditionally, automatic metrics such as accuracy have been used to evaluate models based on the \emph{quality} of their direct answers to specific questions.
However, as \emph{interactions} between humans and LLMs grow more complex and nuanced, these traditional metrics often fail to capture the full spectrum of a model's capabilities (e.g. helpfulness and fluency), particularly in interactive QA settings \citep{liu2016, deriu2021survey}, whereas interactions are crucial for user experience and system effectiveness, yet remains overlooked in traditional evaluation paradigms.
Recent works such as \citep{lee2023evaluating} evaluated human-LLM interactions, a process that involves human participation and annotation. Although human evaluations for these interactions provide a closer approximation to real-world use cases, this approach is significantly costly and time-consuming. Recognizing the need for automatic evaluation, works like G-Eval \citep{liu2023gpteval}, LLM-Eval \citep{lin2023llm}, FaithScore \citep{jing2023faithscore}, and PRD \citep{li2024prd} proposed to automate the assessment of non-interactive LLM responses using LLMs as evaluators.

Drawing insights from (a) using LLM for automatic evaluation and (b) literature of LLM-agents research \cite{wang2024survey, deshpande2023toxicity}, we propose \ours\ framework to auto-evaluate the performance of IQA models (i.e. LLMs) with LLM-based Evaluation Agent~(LEA) to simulate and then evaluate interactions. By additionally incorporating personas, our experiments on a well-annotated dataset show that our methods align well with human judgments and provide a more comprehensive evaluation of LLMs in interactive settings than traditional metrics. 
Finally, we benchmark recent LLMs with new complex and ambiguous questions, and demonstrate that the accuracy of answers does not always transfer to the corresponding ranking of models on their capability of achieving good human-model interactions. See an overview of our system in Figure \ref{fig:sample}.

Our contributions are as follows:
\begin{itemize}[leftmargin=10pt, itemsep=5pt]
   \vspace{-1mm}
    \item We propose the first LLM agent-based automatic evaluation framework \ours\, designed specifically to generate and then evaluate interactions in IQA. Our results demonstrate a strong correlation with human evaluations.
    \vspace{-1mm}
    \vspace{-1mm}
    \item We propose persona-based LLM evaluation agents to better assess how models adapt to different user preferences and interaction styles.
    \vspace{-1mm}
    \item We experiment with \ours\ framework to benchmark the most recent LLMs on the IQA task, demonstrating the strength and general effectiveness of our framework on fully automated evaluation. 
    
\end{itemize}

\section{Related Work}

\paragraph{Evaluating Interactions}

Traditional methods for human-model dialogue evaluation have often been centered around single-turn pairwise evaluation    \citep{vinyals2015neural, li2016persona}. Some methods with multi-turn Likert scores emerge   \citep{venkatesh2017evaluating,zhang2018personalizing, see2019goodconversation} but require time-consuming and costly collection of human-model conversations. In response,   \cite{ghandeharioun2019} suggests a self-play scenario where a dialog system engages in conversation with itself, employing a set of proxy measures for evaluation.

Acute-eval \citep{li2019acute} and LLM-Eval \citep{lin2023llm} have advanced multi-turn evaluation frameworks, reflecting the increasing demand for sophisticated techniques that holistically capture human-AI interactions. Further studies in specific interaction domains like task completion \citep{liu2023}, code generation \citep{yang2023}, and collaborative problem-solving \citep{lee2023evaluating, huang2023, fu2023improving} emphasize the need for evaluations that consider both environmental and human elements \citep{wang2023mint}.  Our work differs from these methods by introducing an automated approach that emphasizes interaction quality and significantly reduces the reliance on human annotations.

\vspace{-2mm}
\paragraph{LLM-based Agent for Simulation}
Recently, LLMs rises to demonstrate human-like intelligence, as evidenced in various studies \citep{xiao2023far, rao_can_2023, li_does_2023, jiang_personallm_2023, srivastava2023beyond, brown2020language, touvron_llama_2023, chowdhery_palm_2022}. The integration of LLMs into agents that simulate complex human behaviors and social interactions is an area of growing research interest \citep{maes_artificial_1995, wooldridge_intelligent_1995, xi_rise_2023}. For example, \citet{park_social_2022} and  \citet{gao_s3_2023} employ these agents to simulate human emotions, attitudes, and behaviors in social networks, while \citet{Park2023GenerativeAgents} leverages in-context learning to simulate human behaviors in sandbox world. Moreover, \citet{argyle_out_2022} utilizes ``algorithmic bias" inherent in GPT-3 to reflect the response patterns of different human subgroups. \citet{horton2023large} utilize LLMs in experimental setups of behavioral economics experiments to facilitate pilot studies. Additionally, \citet{hamalainen_evaluating_2023} and \citet{wang_when_2023} investigate LLM-based agents in recommender systems to simulate and collect data on user behavior. These studies show the broad applicability and potential of LLM-based agents in simulating human behaviors and interactions across diverse applications. Our work utilizes LLM-based evaluation agents (LEAs) to fully automate interactive quality assessments, handling both interaction generation and evaluation, to enhance the evaluation of IQA models in realistic scenarios.

\vspace{-2mm}
\paragraph{Personas in NLP}
Personas are constructed profile prompts that represent key traits of a group of users, as defined in the HCI field, reflecting their characteristics, behaviors, and goals to guide the design of technologies that are well-suited to user needs \citep{cooper2014face}. This approach enhances relevance and personalization in NLP applications \citep{nargund2022persona, bamman2015people, sheng2021revealing, zhong2020towards}, offering significant potential for customizing engagement and improving the effectiveness of conversational agents \citep{li2016persona, zhang2018personalizing, chan2019modeling, madotto2019personalizing, zheng2019pretraining}. \citet{li2016persona} introduces a persona-based neural conversation model to enhance dialogue personalization and coherence. \citet{zhang2018personalizing} develops personalized dialogue agents that incorporate user-specific details to enhance interaction.
In our work, persona settings enable our framework to tailor interactions and assessments, aligning more closely with the specific characteristics and preferences of different user groups.

\section{\ours: Evaluating Interactive Question Answering (IQA)}

In this section, we introduce our \ours\ framework for automatically evaluating Interaction Question Answering Models (IQA models) with \textbf{LLM-based Evaluation Agents (LEA)}.
LEAs are used to simulate humans in the following two stages:
(1) generating interactions with IQA models; and 
(2) evaluating interactions.
Lastly, we discuss the use of personas to for LEAs.

\subsection{Interaction Generation  with LEA (Stage 1)}

Inspired by peer discussions~\cite{lee2023evaluating, wang2024survey}, we prompt LEAs to simulate human behaviors for effective interaction generation with IQA models. The structured prompt includes three key components: (1) a role description; (2) a task description; and (3) instructions for the discussion.

\textbf{Role Description} outlines the people that the LEA model will simulate during interactions. For example, the description for a standard persona could be:
{\ttfamily
\small
\text{You are mimicking a human.}
}

\textbf{Task Description} briefly describes the action that LEA model needs to perform in the task. For example, in a multi-choice question answering task, the prompt could be structured as follows:
{\ttfamily
\small
You are trying to choose the correct answer for the given question.}
Both role and task descriptions can be adjusted based on the persona, as discussed in Section \ref{sec:persona_method}.

\textbf{Discussion instruction} guides LEAs on their subsequent steps by providing detailed descriptions to facilitate progress in interactions. It comprises two essential components: (1) the actions to take; and (2) the detailed procedures to follow. For example, in a question answering task, the prompt specifies:
{\ttfamily
\small
You can ask an assistant questions for help. Please ask sub-questions 
to approach answers. In each turn, please only ask one sub-question 
to interact with the assistant.

In the sub-question, please include all necessary information in the 
original question, such as the question and all options. If you know 
the answer, please output "So, the answer is: A, B, C, or D".
}

At the start of an interaction, the LEA receives a system prompt that includes all three components above, along with the specific question to be addressed.
As the LEA interacts with the IQA model, it generates sub-questions to request clarification of unknown entities, definitions, or particular aspects of the original question.
Then, the IQA Model takes the questions as the input and output responses.
After receiving responses, the LEA continues to pose further questions until it determines the final answer. The full prompt structure and interaction details are provided in Appendix \ref{sec:model-inter-full-prompt}.

\subsection{Interaction Evaluation with LEA (Stage 2)}

Inspired by G-eval \citep{liu2023gpteval}, which demonstrates that evaluations by GPT models align closely with human assessments in NLG tasks, we propose utilizing LEAs for interaction evaluation. LEAs assess interactions generated by LEAs and IQAs in Stage 1. The module takes task details, such as questions or articles, and interactions as the input, and output evaluation scores. The prompt contains three parts: (1) role and task description; (2) metrics definition; and (3) evaluation instruction.

\textbf{Role and task description} instructs LEA to conduct evaluation. The role description acts the same as the role description in the previous stage. 
Moreover, it briefly describes the evaluation task.
The general prompt looks like:
{\ttfamily
\small
You are a helpful and precise evaluator who checks the quality of the AI assistant's responses in interactions.}

\textbf{Metrics definitions} describes the criterion that LEA needs to follow in the evaluation process. They can customized for different tasks. For the question answering task, we add the following prompt to define the ``helpfulness'' metric:
{\ttfamily
\small
\begin{align*}
& \text{Helpfulness (5-point Likert): How helpful was having access to the} \\
& \text{AI Assistant compared to not having access?}
\end{align*}
}
Both the aforementioned parts can be tailored according to the persona that the LEA simulates in Stage 1, as detailed in Section \ref{sec:persona_method}.

\textbf{Evaluation instruction} outlines the specifics of the evaluation task and the required output format. This section may appear as a separate part of the prompt. For example, the instruction within the prompt might be structured as follows:
{\ttfamily
\small
\begin{align*}
& \text{Please evaluate the above interactions between user and AI assistant } \\
& \text{by using the following metrics:} \\
& \text{<Metric definitions>} \\
& \text{Please output each of the above metrics line-by-line.}
\end{align*}
}
Finally, all evaluation scores for metrics are calculated by averaging the results of multiple runs. 
The complete prompt is available in Appendix \ref{sec:meta-eval-full-prompt}, and further details about our implementation can be found in Section \ref{sec:meta_eval}.

\subsection{Assigning the Personas to LEA
}

\label{sec:persona_method}
Both aforementioned evaluation stages typically use a default persona. While this constitutes a somewhat neutral baseline in knowledge, language proficiency, and beliefs to bo baseline, individual users often exhibit diverse personal preferences and characteristics, making a one-size-fits-all evaluation less effective.
Moreover, the persona distribution of the target user group significantly impacts the performance of IQA models in real-world applications. For instance, if 20\% of human users prefer brief interactions and 70\% prefer detailed information, applying a general LEA to simulate this group of persons is likely to result in poor correlation with downstream users.

To better simulate the diversity of the groups of people and provide individualized evaluations, we assign personas to LEAs. This affects prompts of both interaction generation and interaction evaluation processes.

For example, when the LEA is assigned with the ``Critical-Seeker'' persona (definition in \ref{sec:persona-prompt}) for interaction generation, 
we adapt the default role and task description (in Stage 1) to:\\
{\ttfamily \small You prefer interactions rich in critical information. You need help from an assistant and try to get critical information from it to answer the following questions.}
For interaction evaluation, the default role and task description prompt changes to {\ttfamily \small The AI Assistant should provide straightforward, simple, and concise answers to aid users in deducing solutions.}
Additionally, the definition of metrics is also adjusted to align with this persona, with further details available in Appendix \ref{sec:persona-prompt}.

\section{Meta-Evaluation of \ours\ Framework}
\label{sec:meta_eval}

To measure how our framework provides trustworthy IQA evaluations that align with human preferences, we conduct meta-evaluations experiments and report correlation scores. 

\vspace{-2mm}
\subsection{Experiment Settings}
\label{sec:experiment-settings}
\paragraph{Dataset and Evaluation Metrics}
We apply our evaluation method on the annotated dataset from the study by \citet{lee2023evaluating}. This dataset consists of 3641 interactions from 331 annotators. Questions in the dataset are multi-choice and are derived from the MMLU dataset \citep{hendrycks2020} (example question in Figure~\ref{fig:sample}).
The construction of MMLU requires each worker to participate in 11 random conversations with one of the following three IQA models: \gpttextdavinci (\texttt{text-davinci-001}), \gpttextbabbage (\texttt{text-babbage-001}), and \gptdavinci (\texttt{davinci-001}). 
At the end of conversations, fluency and helpfulness scores are annotated by annotators. The number of queries and accuracy for each IQA model can be easily deduced from annotations.
In this work, We adjust the four metrics to evaluate generated interactions:
\vspace{-1mm}
\begin{itemize}[leftmargin=15pt, itemsep=5pt]
    \item \textbf{Fluency (5-point Likert)}: How clear (or fluent) were the responses from the AI Assistant?
    \vspace{-1mm}
    \item \textbf{Helpfulness (5-point Likert)}: Independent of its fluency, how helpful was having access to the AI Assistant compared to not having access?
    \vspace{-1mm}
    \item \textbf{Number of Queries}: Counts the number of interaction turns in the conversation. This metric helps assess the efficiency of the AI in resolving queries within a minimal number of interactions.
    \vspace{-5mm}
    \item \textbf{Accuracy}: Quantifies how accurately the AI's responses match the golden answers. This is critical for evaluating the correctness of the AI's knowledge and its application in practical scenarios.
\end{itemize}

\paragraph{LEA Models}
To evaluate the effectiveness of \ours\ framework, we experiment with different LEA models on the above-mentioned three LLMs IQA models in MMLU.
For LEA that conducts both interaction generation and interaction evaluation,
we use \chatgpt (\texttt{GPT-3.5-turbo-1106}), \gptfour (\texttt{GPT-4-1106-preview}), \claude (\texttt{Claude-1}).

\paragraph{Evaluation of \ours\ Framework}
We report \textbf{Pearson correlations} as the measrue of agreement between the Human evaluations and LEA evaluations of IQA models.

\subsection{Experiment Results}

According to Table \ref{tab:model-interaction-correlation}, all models, including \gptfour, \gptthree, and \claude, show high correlation with human evaluations.
\gptfour aligns most closely with human judgments in both ``Helpfulness'' and ``Fluency" metrics and the highest overall correlation score. 
This indicates that these models are capable of effectively performing \ours\ framework as LEA models.

\begin{table*}[t]
\caption{\ours\ evaluation results of IQA models (TDA: \gpttextdavinci; TB: \gpttextbabbage; DA: \gptdavinci).
\textbf{Bold numbers} indicate they are the most close to human results.
The empty set symbol (Ø) indicates the number cannot be calculated due to the model's inability to follow instructions and produce a gradable answer.
}
\resizebox{\textwidth}{!}{%
\begin{tabular}{l|ccc|ccc|ccc|ccc}
\toprule
         & \multicolumn{3}{c|}{\textbf{Helpfulness}}     & \multicolumn{3}{c|}{\textbf{Fluency}}         & \multicolumn{3}{c|}{\textbf{\# Queries}} & \multicolumn{3}{c}{\textbf{Accuracy}} \\
Evaluator       & TDA            & TB            & DA             & TDA            & TB            & DA             & TDA           & TB          & DA           & TDA             & TB             & DA               \\
\midrule
Human    & 4.60          & 3.84          & 3.52          & 4.35          & 3.84          & 3.22          & 1.78         & 2.57        & 2.66        & 0.69          & 0.52          & 0.48                    \\
\midrule
\ours-\gptfour    & 3.67          & 2.30          & 2.10          & 4.77          & \textbf{3.87} & 3.03          & \textbf{1.57}         & 2.27        & \textbf{2.37}        & 0.87         & 0.83         & 0.67                     \\
\ours-\claude & 4.13          & 3.03          & 3.00          & \textbf{4.47} & 3.47          & \textbf{3.23} & 2.20         & \textbf{2.67}        & 2.07        & \textbf{0.67} & \textbf{0.53} & 0.57                    \\
\ours-\gptthree  & \textbf{4.30} & \textbf{3.87} & \textbf{3.93} & \textbf{4.47} & 3.67          & 3.97          & \textbf{1.57}         & 1.77        & 2.00        & 0.63          & 0.47          & \textbf{0.53}            \\
\bottomrule
\end{tabular}%
}
\vspace{-4mm}
\label{tab:model-interaction}
\end{table*}

\begin{wraptable}{R}{0.6\textwidth}
\centering
\small
\vspace{-9mm}
\caption{Pearson Correlation ($\rho$) between \ours\ evaluations and human judgments.}
\begin{tabular}{l|ccc}
\toprule
                    & \textbf{Helpfulness} & \textbf{Fluency} & \textbf{Overall} \\ \midrule
\ours-\gptfour      & \textbf{0.652}        & \textbf{0.591}    & \textbf{0.613}            \\
\ours-\claude       & 0.640                 & 0.552             & 0.551            \\
\ours-\gptthree    & 0.621                & 0.523            & 0.510   \\
\bottomrule
\end{tabular}%
\vspace{-8mm}
\label{tab:model-interaction-correlation}
\end{wraptable}

In Table \ref{tab:model-interaction}, \chatgpt scores closest to human judgments, particularly in the "Helpfulness" metric. Conversely, \gptfour and \claude score lower on "Helpfulness" than human evaluations, because they tend to produce inaccurate and repetitive responses that lack coherence and do not directly address user queries, as indicated by their generated explanations. 
IQA-Eval scores on “Fluency” are close and highly correlated to human judgments. Both scores given by humans and LEA models show that IQA models provide fluent outputs.
Furthermore, according to the "\# Queries" metric, most models conclude conversations more quickly than humans, except for \claude, which requires more turns, potentially due to its non-OpenAI origins that it needs more turns to adapt the conversational style and understand responses.
Notably, \gptfour achieves the highest accuracy among all models.
Moreover, we consider the impact of self-enhancement bias and conduct more experiments. Details are in Section \ref{sec:debias}.

\vspace{-2mm}
\paragraph{Analysis of LEA for Stage 2 (Evaluating Interactions)}

For Stage 2 itself, we measure the LEA's capability of evaluating interactions, based on real human-generated interactions from Stage 1.

\begin{figure}[h]
    \centering
    \vspace{-4mm}
    \begin{subfigure}[b]{0.49\textwidth} 
        \centering
        \includegraphics[width=\textwidth]{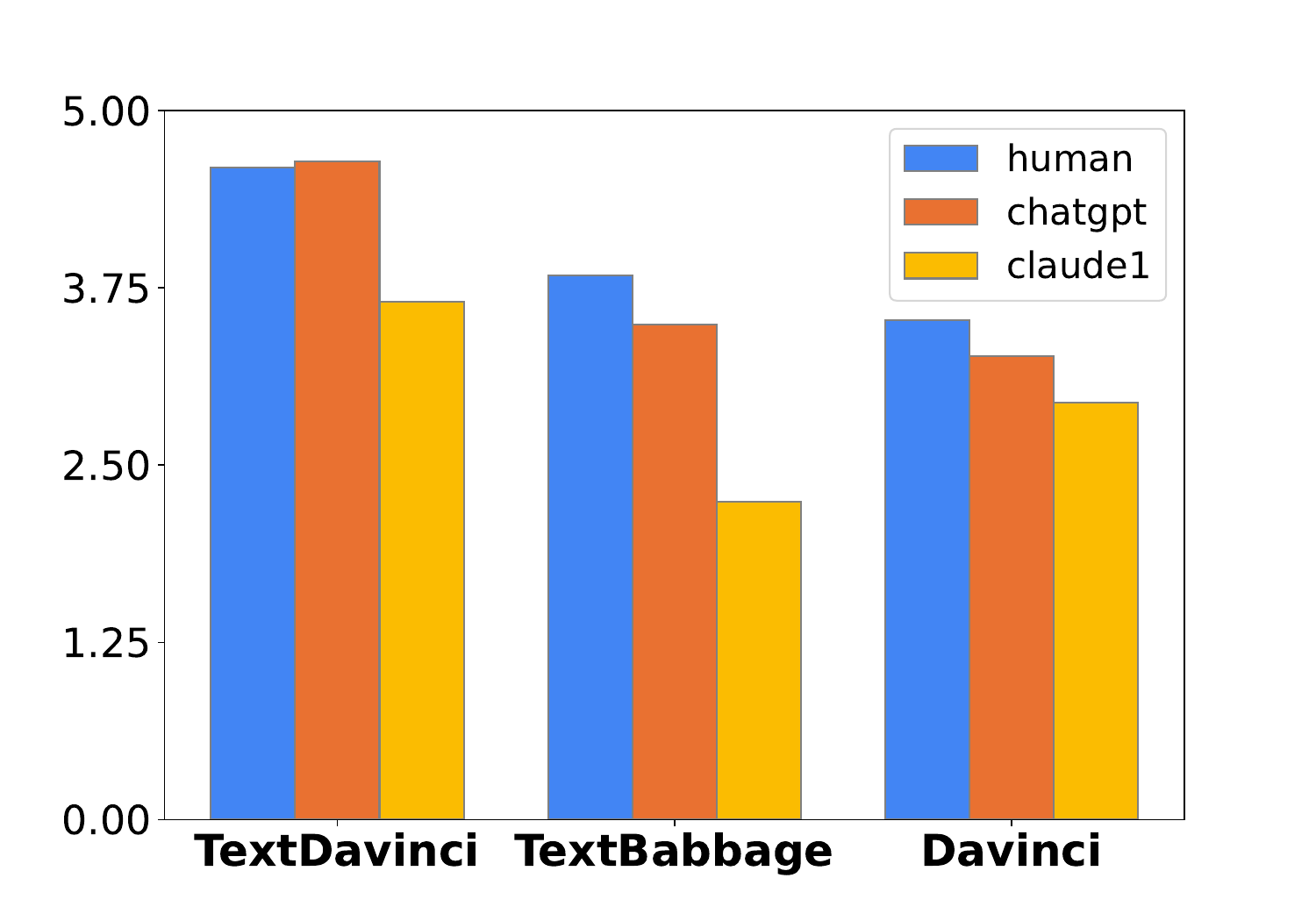}
        \vspace{-0.7cm}
        \caption{Helpfulness}
        \label{fig:helpfulness}
    \end{subfigure}
    \hfill
    \begin{subfigure}[b]{0.49\textwidth} 
        \centering
        \includegraphics[width=\textwidth]{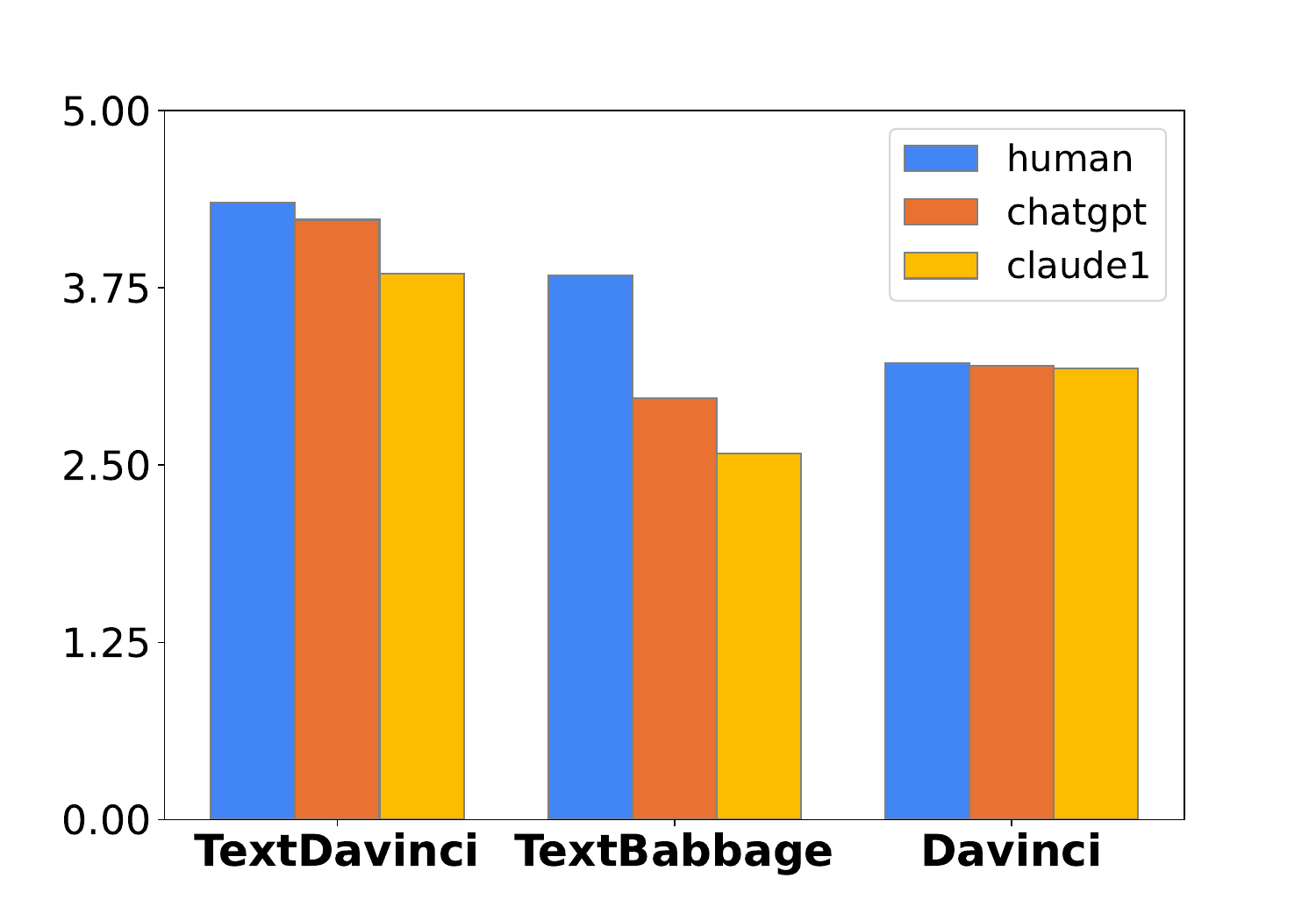}
        \vspace{-0.7cm}
        \caption{Fluency}
        \label{fig:fluency}
    \end{subfigure}
    \caption{Interaction evaluation results evaluated by human and two LEA models on interactions between real human and IQA Models. All scores are on a scale of 5.
    }
    \vspace{-2mm}
    \label{fig:third-party-comparison}
\end{figure}

Results are in shown Figures \ref{fig:helpfulness} and \ref{fig:fluency}.
The Pearson correlation coefficients for fluency and helpfulness between human judgments and LEA evaluations show distinct patterns. 
\chatgpt demonstrates a stronger correlation with human ratings, recording a correlation score of 0.424 for fluency and a correlation score of 0.306 for helpfulness. 
In contrast, \claude shows slightly lower correlations (0.281 for fluency and 0.287 for helpfulness). 
This shows that \chatgpt aligns better with human compared to \claude in these specific metrics.

Moreover, for interactions between humans and IQA models (Figure \ref{fig:third-party-comparison}), LEA evaluations moderately correlate with human evaluations. However, for interactions between LEAs and IQA models (Table \ref{tab:model-interaction-correlation}), which is the main focus of our paper, LEA evaluations highly correlate (around 0.6) with human evaluations. This indicates that LEA models are helpful when participating in the whole evaluation process, including both the interaction and evaluation. However, when evaluating interactions between humans and IQA models, LEAs focus differently from humans, which causes moderate correlations.

Thus, the results show that (1) both models' evaluations moderately correlate with human evaluation; (2) \chatgpt's evaluation is closer and related to human evaluation than \claude's.

\subsection{Further Analysis for Free-form Feedback}

Apart from the metrics above, we also prompt LLM-based Evaluation Agent (LEA) to explain the reason for generating scores in the free-form text format.
We find that: 1) \textbf{\chatgpt Generates more human-like, positive reviews}:
\chatgpt evaluations are generally more positive, frequently using terms like ``helpful'', ``relevant'', and ``useful'' -- words not always noted by workers in their annotations. 
Despite this, \chatgpt often identifies similar issues as human raters, such as the provision of irrelevant and repetitive information.
Overall, \chatgpt's assessments align well with human evaluations;
2) \textbf{\claude flags more Issues}:
\claude is more strict and critical to IQA models in interactions. In the free-text feedback, \claude tends to highlight more issues with model responses rather than acknowledging positive aspects, especially for \gptdavinci.
For one question, after interacting with the IQA model/assistant (\chatgpt in this case), human and two LEAs provide the following feedback:
{\ttfamily
\small
\begin{align*}
\parbox{\textwidth}{
\textbf{Human}: When rephrasing questions well, the answers could be found in the AI's response.\\
\textbf{\chatgpt:} The AI assistant was helpful in providing relevant information, but there were issues with the accuracy.\\
\textbf{\claude:} The AI assistant's responses were not very helpful. \textit{The responses were often vague, repetitive, or did not directly answer the question.}
}
\end{align*}}

\section{
Effect of Assigning Persona to LLM Evaluation Agent (LEA)
}

\subsection{Persona Definitions}
\label{sec:persona:desc}
As discussed in Section \ref{sec:persona_method}, we assign personas to LEAs to simulate different groups of humans for diverse human alignments. We investigate assigning the following personas to LEAs, which are defined based on the crowdworker survey results in \citet{lee2023evaluating}.
\vspace{-1mm}
\begin{itemize}[leftmargin=0pt,itemsep=1pt]
    \item \textbf{Expert:} knowledgeable; quickly learns new concepts and applies them in the reasoning process to answer questions.
    \item \textbf{Critical-Thinker:} people who prefer critical information rather than redundant or detailed responses.
    \item \textbf{Adaptability-Seeker:} people who prefer assistants can understand their questions even if they are not precise.
    \item \textbf{Clarity-Seeker:} people who prefer clear explanations from assistants.
\end{itemize}

Based on the survey results on the distributions of personas of workers~\cite{lee2023evaluating}, for each persona $P$, we split the crowdworkers into two groups: ``persons with persona $P$'', and ``normal persons without specific persona $P$''. 
For the first group, we initialize the role prompt in Section \ref{sec:persona_method}
and use it for the LEAs.
For the second group, we utilize default prompting for LEAs. The LEA model in this section is ChatGPT(\texttt{GPT-3.5-turbo-1106}).

\vspace{-2mm}

\subsection{Experimental Results}

\vspace{-1mm}

\begin{table}[t]
\centering \small
\caption{
\ours\ evaluation results of IQA models (TDA: \gpttextdavinci; TB: \gpttextbabbage; DA: \gptdavinci).
LEAs, based on \gptthree, are assigned specific personas when representing specific groups of workers.
}
\begin{tabular}{l|ccc|ccc}
\toprule
              & \multicolumn{3}{c|}{\textbf{\# Queries}} & \multicolumn{3}{c}{\textbf{Accuracy}}  \\
Evaluator              & TDA   & TB   & TD                          & TDA   & TB   & TD                      \\ 
\midrule
Human         & 1.78 & 2.57 & 2.66                       & 0.69 & 0.52 & 0.48                  \\
\ours\ & 1.57 & 1.77 & 2.00                       & 0.63 & 0.47 & 0.53                  \\ 
\midrule
\ours\ (Expert)   & 1.20 & 1.49 & 2.20                       & \textbf{0.73} & 0.56 & 0.53                  \\ 
\ours\ (Critical-Thinker)  & 1.55 & 1.80 & 1.99                       & 0.68 & 0.54 & 0.55                  \\ 
\ours\ (Adaptability-Seeker)  & 1.50 & 1.75 & 2.10                       & 0.66 & 0.52 & 0.55                  \\ 
\ours\ (Clarity-Seeker)  & 1.64 & 2.10 & 2.34                       & 0.63 & \textbf{0.57} & \textbf{0.57}                   \\
\bottomrule
\end{tabular}
\vspace{1mm}
\label{tab:persona}
\caption{
\ours\ evaluation results (helpfulness and fluency) of IQA models.
Correlations are between the LEA evaluation in each row and human evaluations.
}
\resizebox{\textwidth}{!}{%
\begin{tabular}{l|cccc|cccc|c}
\toprule
   & \multicolumn{4}{c|}{\textbf{Helpfulness}} & \multicolumn{4}{c|}{\textbf{Fluency}} & Overall  \\
Evaluator              & TDA   & TB   & TD    & $\rho$                                      & TDA   & TB   & TD    & $\rho$                                  & $\rho$     \\ 
\midrule
Human                  & 4.60 & 3.84 & 3.52 & -                                         & 4.35 & 3.84 & 3.22 & -                                     & -        \\
\ours\                 & 4.30 \scriptsize(±0.06) & 3.87 \scriptsize(±0.11) & 3.93 \scriptsize(±0.13) & 0.621                                     & 4.47 \scriptsize(±0.05) & 3.67 \scriptsize(±0.08) & 3.97 \scriptsize(±0.06) & 0.523                                 & 0.510   \\ 
\midrule
\ours\ (Expert)        & 4.17 \scriptsize(±0.08) & 3.08 \scriptsize(±0.09) & 3.12 \scriptsize(±0.11) & 0.756                                     & 4.47 \scriptsize(±0.02) & 3.84 \scriptsize(±0.04) & 3.40 \scriptsize(±0.04) & 0.787                                 & 0.670   \\ 
\ours\ (Critical-Thinker) & 4.44 \scriptsize(±0.08) & 4.02 \scriptsize(±0.13) & 4.08 \scriptsize(±0.17) & 0.711                                  & 4.64 \scriptsize(±0.06) & 3.97 \scriptsize(±0.08) & 4.10 \scriptsize(±0.08) & 0.624                                 & 0.634    \\ 
\ours\ (Adaptability-Seeker) & 4.24 \scriptsize(±0.05) & 3.67 \scriptsize(±0.11) & 3.75 \scriptsize(±0.11) & 0.713                               & 4.52 \scriptsize(±0.08) & 3.84 \scriptsize(±0.07) & 3.84 \scriptsize(±0.09) & 0.637                                 & 0.650   \\ 
\ours\ (Clarity-Seeker) & 4.45 \scriptsize(±0.07) & 3.77 \scriptsize(±0.15) & 3.80 \scriptsize(±0.12) & 0.747                                    & 4.60 \scriptsize(±0.04) & 3.85 \scriptsize(±0.04) & 3.94 \scriptsize(±0.06) & 0.676                                 & 0.690   \\
\bottomrule
\end{tabular}}
\vspace{1mm}
\label{tab:persona-cor}
\vspace{-7mm}
\end{table}

In Table \ref{tab:persona}, we show the results of model (with personas) evaluations.
To accurately simulate persona distribution, each interaction is executed multiple times, with different personas (including the standard one) assigned to LEA based on their distribution proportions. This method ensures that each persona's influence and characteristics are proportionally represented in the simulation, reflecting their respective prevalence within the overall distribution. The final score for a persona is an average of all experiment results.

The ``Expert'' persona decrease LEA query counts as ``Expert'' already possesses relevant knowledge and only needs key explanations.
The ``Clarity-Seeker'' requires the most interaction turns among all personas for comprehension, but achieves the highest accuracy with \gpttextbabbage and \gptdavinci through detailed understanding of questions. 

``Critic-Thinker'' and ``Adapability-Seeker'' in Tables \ref{tab:persona} and \ref{tab:persona-cor} rarely surpass upon the standard persona's human-preference aligment. We hypothesize that these personas are less reflected within the overall distributions of human preferences. 
In Table \ref{tab:persona}, the varying ``\# Queries'' across personas reveals their significant influence on LEA interaction strategies. Accuracy remains consistent after adding personas, showing no performance degradation.

Together, these results indicate that assigning specific personas steer LEAs to perform \ours\ in a more fine-grained and human-aligned way.

It is worth nothing that our analysis of persona reassignment shows \ours\ is sensitive to incorrect assignments (see Appendix \ref{sec:persona-distribution}).
Moreover, further analyses about bias evaluation, as well as measuring complementary metrics like offensiveness, are in Appendix \ref{sec:ethics}.

\vspace{-2mm}

\section{Benchmarking LLMs with \ours\ on more Types of Questions}
\label{sec:eval_full}

\subsection{Datasets}
To evaluate the robustness and generalizability of our evaluation framework, we conduct benchmarking across different models on two distinct question answering datasets, each offering unique challenges and complexities requiring advanced reasoning.
\textbf{AmbigQA}~\cite{min-etal-2020-ambigqa} is a collection of 14,042 annotated questions sourced from the NQ-OPEN benchmarks \cite{10.1162/tacl_a_00276}, an open-domain QA dataset. It focuses on questions with inherent ambiguities, reflecting the complexity encountered in real-world queries. These ambiguities often involve diverse aspects such as events, entity references, and answer types, resulting in multiple plausible answers for each question.
\textbf{HotpotQA}~\cite{yang2018hotpotqa} comprises 113,000 question-answer pairs sourced from Wikipedia, which require multi-hop reasoning spanning multiple documents. It contains a rich array of intricate questions that demand the synthesis of information from various texts to determine accurate answers.
In this benchmark, we select 500 questions from each dataset to form a dataset containing 1,000 complex multi-hop and ambiguous questions.

\subsection{LLMs to Benchmark}

\begin{table}[ht]
\caption{
\ours\ benchmarking results on HotpotQA and AmbigQA datasets. 
}
\resizebox{\columnwidth}{!}{%
\begin{tabular}{l|cccc|cccc}
\toprule
\multirow{2}{*}{\textbf{IQA Models}} & \multicolumn{4}{c|}{\textbf{HotpotQA}}                                            & \multicolumn{4}{c}{\textbf{AmbigQA}}                                              \\
                                     & \textbf{Helpfulness $\uparrow$} & \textbf{Fluency$\uparrow$} & \textbf{\# Queries$\downarrow$} & \textbf{Accuracy$\uparrow$} & \textbf{Helpfulness} & \textbf{Fluency} & \textbf{\# Queries} & \textbf{Accuracy} \\ 
\midrule
\gpttextdavinci                          & 4.72                 & 4.87             & 1.22                & 0.45              & -            & -       & -          & -        \\
\gpttextbabbage                          & 4.70                 & 4.88             & 1.74                & 0.37              & -            & -       & -          & -        \\
\gptdavinci                              & 4.27                 & 4.52             & 1.68                & 0.32              & -            & -       & -          & -        \\
\midrule
\gptthree                                & 4.72                 & 4.95             & 1.49                & 0.63              & 4.91                 & 4.97             & 1.89                & 0.60              \\
\gptfour                                 & 4.78                 & 4.96             & 1.12                & 0.66              & 4.89                 & 4.95             & 1.06                & 0.72              \\
\claude                                  & 4.82                 & 4.99             & 1.26                & 0.58              & 4.89                 & 4.94             & 1.36                & 0.62              \\
\llama                                   & 4.70                 & 4.95             & 1.32                & 0.55              & 4.96                 & 4.94             & 1.79                & 0.52              \\
\zephyr                                  & 4.64                 & 4.88             & 1.01                & 0.40              & 4.38                 & 4.66             & 1.03                & 0.45             \\
\bottomrule
\end{tabular}%
}
\label{tab:benchmark}
\vspace{-3mm}
\end{table}

Apart from \gpttextdavinci, \gpttextbabbage and \gptdavinci we benchmark more LLMs: \gptthree, \gptfour, \claude, \llama and \zephyr.
The checkpoints for \llama and \zephyr are Llama-2-7B and Zephyr-alpha, respectively.
\gptthree is used as LEA in our experiments.

\subsection{Benchmarking Results}
The IQA evaluation benchmarks are presented in Table \ref{tab:benchmark}. We divide IQA Models into two categories: weak IQA Models (\gpttextdavinci, \gpttextbabbage, and \gptdavinci) and strong IQA Models (\gptthree, \gptfour, \claude, \llama, and \zephyr).
Weak IQA Models can assist the LEA with answering HotpotQA questions, but due to their knowledge limitations, they cannot help much with AmbigQA questions.
\zephyr achieves the lowest performance compared to other strong IQA Models. 
On the HotpotQA dataset, \zephyr's accuracy performance is only comparable to the strongest one among weak IQA Models, \gpttextdavinci.

Most ``Helpfulness'' and ``Fluency'' scores are high (exceeding 3 out of 5), especially for strong IQA Models like \gptfour.
For the ``\# of queries'', it is uncommon for interactions to extend beyond two turns. As on HotpotQA, most interactions conclude at the beginning of the second turn, as IQA models have effectively guided users to reach the answers.
For AmbigQA, some conversations last longer whereas LEA spends additional turns on clarifying ambiguous entities before approaching the final answer.
Additional benchmarking results on the Natural Question dataset are in Appendix \ref{sec:benchmarking-nq}.

\subsection{Self-Enhancement Bias}
\label{sec:debias}
LLMs are shown to demonstrate self-favouring behaviours \citep{panickssery2024llmevaluatorsrecognizefavor}, and no verified or accessible mitigations to this issue exist to the best of our knowledge. This issue is particularly concerning when the LEA models evaluating the IQA models share the same underlying model. In this section, we discuss our two of our attempts to assess the effects of this bias.

Following \cite{zheng2023judging} and \cite{furniturewala2024thinking}, we included some empirically useful debiasing instructions as follows:
{\ttfamily
\small
\begin{align*}
\parbox{\textwidth}{
Please act as an impartial and unbiased judge. In your evaluation, please be objective and do not include any bias or your preference.
}
\end{align*}
}

\begin{table}[]
\caption{Comparison between new prompts and our prompts used in Table \ref{tab:model-interaction}. The new prompts are more complex and include effective debiasing instructions.}
\resizebox{\columnwidth}{!}{%
\begin{tabular}{l|lll|lll|lll}
\toprule
\textbf{}                     & \multicolumn{3}{c|}{\textbf{Helpfulness}} & \multicolumn{3}{c|}{\textbf{Fluency}} & \multicolumn{3}{c}{\textbf{Accuracy}} \\
LEA models                    & TDA          & TB           & DA          & TDA         & TB         & DA         & TDA         & TB         & DA         \\
\midrule
Human                         & 4.60         & 3.84         & 3.52        & 4.35        & 3.84       & 3.22       & 0.69        & 0.52       & 0.48       \\
IQA-EVAL-GPT4 (Our Prompts)   & 3.67         & 2.30         & 2.10        & 4.77        & 3.87       & 3.03       & 0.87        & 0.83       & 0.67       \\
IQA-EVAL-GPT4 (New Prompts)   & 3.50         & 2.23         & 2.10        & 4.40        & 4.07       & 3.53       & 0.87        & 0.83       & 0.67       \\
\bottomrule
\end{tabular}%
}
\label{tab:debias}
\end{table}

In Table \ref{tab:benchmark}, scores on the row of \gptthree is vulnerable to self-enhancement bias. However, with the above debiasing prompt, in Table \ref{tab:debias-gpt35} shows that the results of new prompts are highly similar to the original prompts.
Similarly, Table \ref{tab:debias} shows that the results of modified and original prompts are differ only lightly. 
\begin{table}[]
\centering
\caption{
Comparison between new prompts and our prompts used in Table \ref{tab:benchmark} on benchmarking LLMs with \ours\ .
}
\resizebox{0.9\columnwidth}{!}{%
\begin{tabular}{l|cccc}
\toprule
\textbf{LEA models}           & \textbf{Helpfulness} & \textbf{Fluency} & \textbf{\# Queries} & \textbf{Accuracy} \\
\midrule
IQA-EVAL-GPT3.5 (Our Prompts) & 4.72                                     & 4.95                                 & 1.49                                   & 0.63                                  \\
IQA-EVAL-GPT3.5 (New Prompts) & 4.68                                     & 4.91                                 & 1.35                                   & 0.60                                  \\
\bottomrule
\end{tabular}%
}
\label{tab:debias-gpt35}
\vspace{-3mm}
\end{table}

We also designed a second methodology to mitigate self-enhancement bias. In this experiment, multiple LEA models evaluate the performance of IQA models during each interaction (“Multi-perspective”). In other words, we introduced third-party evaluations, where various LEA models assess the IQA models' performance instead of relying solely on the LEA model itself involved in the interaction. After evaluation, we use the average score from all LEA models as the final score. The results of IQA-Eval-Multi-perspective look as in Table \ref{tab:multi-perspective}. The correlations between IQA-Eval-Multi-perspective and human evaluations are in Table \ref{tab:multi-perspective-corr}.

We believe that that the self-preference bias has limited impact on IQA-Eval.
\begin{table}[]
\caption{
IQA-EVAL-Multi-Perspective Results of IQA Models.
MP indicates ``Multi-Perspective''.
Bold numbers indicate they are the closest to human results.
}
\resizebox{\columnwidth}{!}{%
\begin{tabular}{l|ccc|ccc|ccc|ccc}
\toprule
                                  & \multicolumn{3}{c|}{Helpfulness}               & \multicolumn{3}{c|}{Fluency}                   & \multicolumn{3}{c|}{\# Queries}                 & \multicolumn{3}{c}{Accuracy}                  \\
LEA models                        & TDA           & TB            & DA            & TDA           & TB            & DA            & TDA           & TB            & DA            & TDA           & TB            & DA            \\
\midrule
Human                             & 4.60          & 3.84          & 3.52          & 4.35          & 3.84          & 3.22          & 1.78          & 2.57          & 2.66          & 0.69          & 0.52          & 0.48          \\
\midrule
IQA-EVAL-GPT4-MP   & \textbf{4.32} & \textbf{3.70} & \textbf{3.53} & 4.57          & \textbf{3.74} & 3.68          & \textbf{1.57} & 2.27          & \textbf{2.37} & 0.87          & 0.83          & 0.67          \\
IQA-EVAL-Claude-MP & 3.96          & 3.13          & 3.10          & \textbf{4.29} & 3.51          & \textbf{3.22} & 2.20          & \textbf{2.67} & 2.07          & \textbf{0.67} & \textbf{0.53} & 0.57          \\
IQA-EVAL-GPT3.5-MP & 3.98          & 3.23          & 3.04          & \textbf{4.41} & 3.67          & 3.59          & \textbf{1.57} & 1.77          & 2.00          & 0.63          & 0.47          & \textbf{0.53}  \\
\bottomrule
\end{tabular}%
}
\label{tab:multi-perspective}
\vspace{-5mm}
\end{table}

\begin{table}[H]
\centering
\caption{Pearson Correlation ($\rho$) between IQA-EVAL-Multi-Persepctive evaluations and human judgments.}
\resizebox{0.6\columnwidth}{!}{%
\begin{tabular}{l|ccc}
\toprule
LEA models & Helpfulness    & Fluency        & Overall        \\
\midrule
IQA-EVAL-\gptfour-MP       & \textbf{0.702} & {0.601} & \textbf{0.624} \\
IQA-EVAL-\claude-MP     & 0.663          & \textbf{0.613}          & 0.602          \\
IQA-EVAL-\gptthree-MP     & 0.641          & 0.552          & 0.533          \\
\bottomrule 
\end{tabular}%
}
\vspace{1mm}
\label{tab:multi-perspective-corr}
\vspace{-3mm}
\end{table}

\subsection{Analysis}

\begin{wraptable}{R}{0.4\columnwidth}
\centering
\small
\vspace{-5mm}
\caption{Accuracy of IQA Models (recent LLMs) on two datasets (Non-interactive setting).}
\resizebox{0.4\columnwidth}{!}{%
\begin{tabular}{l|cc}
\toprule
\textbf{IQA Models} & \textbf{HotpotQA} & \textbf{AmbigQA} \\
\midrule
\gptthree             & 0.43              & 0.62             \\
\gptfour               & \textbf{0.46}              & \textbf{0.63}             \\
\claude              & 0.28              & 0.5              \\
\llama             & 0.24              & 0.29             \\
\zephyr              & 0.25              & 0.31            \\
\bottomrule
\end{tabular}%
}
\label{tab:benchmark-statistic}
\vspace{-9mm}
\end{wraptable}

\paragraph{Stronger IQA models require fewer turns in interactions.}
On the more challenging AmbigQA, the stronger model, \gptfour, typically requires only one turn to assist LEA in solving questions with high accuracy.
In contrast, less capable models like \llama and \gptthree need more turns to clarify ambiguous entities and have lower QA accuracies.
A similar trend is observed on the HotpotQA.

\paragraph{We obtain a similar model ranking with a much lower cost.}
Compared to Chatbot Arena \footnote{\url{https://huggingface.co/spaces/lmsys/chatbot-arena-leaderboard}}, our accuracy-based ranking of IQA Models follows a similar trend: \texttt{\small \gptfour > \claude > \gptthree > \llama > \zephyr}.
In addition, our evaluation method, \ours, is fully automated.
Our method makes it a cost-effective alternative for large-scale evaluations.

\paragraph{Evaluation of interaction performance does not always match Non-Interaction performance.} In interaction evaluations, accuracy on final results is not the only metric to show IQA Models' performance. The quality of intermediate responses is a significant aspect. On both ``helpfulness'' and ``fluency'' metrics, \claude is always the best IQA Model on HotpotQA questions, while on AmbigQA, \llama and \gptthree outperform \gptfour. IQA Model rankings on these two aspects differ from those in Chatbot Arena (non-interaction).

\paragraph{The performance of IQA Models largely affects the final performance.}%

The accuracies in both Table \ref{tab:benchmark} and Table \ref{tab:benchmark-statistic} show a consistent trend. The Pearson correlations of the accuracy between the tables are 0.77 and 0.87 on both datasets, respectively. A strong IQA model, such as \gptfour, can lead the LEA to finish tasks and largely improve the LEA's performance on those tasks. Weak assistants may drag down the LEA's performance, such as the performance of the LEA on both datasets decreases after interacting with \zephyr.

\section{Conclusion}
To conclude, we introduced \ours, a novel approach for evaluating interactive question-answering systems using large language models.
Our methodology achieves automatic interaction generation and evaluation with LEA, and enhances the evaluation process by assigning personas to LEA for better matching diverse groups of people.
We show that our approach aligns closely with real human interactions and judgment, indicating that a scalable, automatic \ours\ process can be achieved.
We providing insights on recent LLM's capability in conducting IQA with \ours\, which would cost \$5,000 for human evaluations.

\section*{Acknowledgement}
We thank the anonymous reviewers for valuable and insightful feedback.
This research is supported in part by the National Science Foundation CAREER Grant IIS-2340435, Amazon Research Award and Cisco Research Award.  Any opinions, findings, and conclusions or recommendations expressed herein are those of the authors and do not necessarily represent the views, either expressed or implied, of the U.S. Government.

\bibliography{custom, colm2024_conference, reference}
\bibliographystyle{abbrvnat}

\appendix
\newpage

\section{Limitations}
\label{limitations}
In conducting this study, certain limitations have influenced our scope and findings. First and foremost, LLMs are shown to demonstrate self-favouring behaviours \citep{panickssery2024llmevaluatorsrecognizefavor}, and no verified or accessible mitigations to this issue exist to the best of our knowledge. We discuss some of our attempts to address this in Section \ref{sec:debias}, but this limitation necessitates future research.

 Our methodology was applied exclusively to multi-choice question-answering tasks due to constraints imposed by the datasets used. 
Moreover, we do not investigate how allowing prompt editing could affect the results. This choice limits the generalizability of our findings across the wider array of question answering formats that exist in both academic research and practical applications. 

We advocate for subsequent research efforts to extend the application of our proposed evaluation strategies to a more diverse set of question answering tasks, beyond the multi-choice format. 

Furthermore, there is a significant opportunity to test these methods with a broader spectrum of LLMs, including those at the cutting edge of the field. Such expansions would not only validate the versatility and robustness of our approaches but also potentially uncover additional insights into the nuances of LLM interaction and performance in varied contexts.
\section{Ethics Statement}

In our study, we meticulously crafted each persona to mitigate bias, ensuring they do not adversely impact the IQA-Eval process, as evidenced by the results presented in the table above. However, our personas represent only a limited range. There exists the potential for negative effects from other personas that may be inadequately designed or deliberately biased to achieve specific outcomes. Thus, we oppose irresponsible persona designs that result in biased evaluation results. The principle of persona definition and design should be thoroughly studied in future works. We hope our work will help facilitate future research into better automatic interaction evaluations aligning with crowds.

\section{Prompts}

Both \ref{sec:meta-eval-full-prompt} and \ref{sec:model-inter-full-prompt} are adapted from \citet{lee2023evaluating}. All the following prompts are from the view of LEA in interactions.

\subsection{Interaction Evaluation Prompt}

\label{sec:meta-eval-full-prompt}

We follow the setting in the data and evaluate conversations worker-wise. In other words, we combine conversations for each worker and send all of them to evaluators. The prompt we send to APIs follows the following format:
{\ttfamily
\begin{align*}
\small
& \text{You are a helpful and precise assistant for checking the quality}\\
& \text{of the AI assistant's responses in interactions.} \\
& \\
& \text{\{Question 1\}} \\
& \text{\{Golden Answer 1\}} \\
& \text{\{Conversation 1\}} \\
& \text{\{User Answer 1\}} \\
& \cdots \\
& \text{\{Question $n$\}} \\
& \text{\{Golden Answer $n$\}} \\
& \text{\{Conversation $n$\}} \\
& \text{\{User Answer $n$\}} \\
& \\
& \text{Please evaluate the above conversations between user and AI assistant } \\
& \text{by using the following metrics:} \\
& \text{Fluency (5-point Likert): How clear (or fluent) were the responses} \\
& \text{from the AI Assistant?} \\
& \text{Helpfulness (5-point Likert): Independent of its fluency, how helpful} \\
& \text{was having access to the AI Assistant compared to not having access?} \\
& \text{Helpfulness (free-form): Why did you find the AI Assistant helpful} \\
& \text{or unhelpful?} \\
& \text{Please output each of the above metrics line-by-line.}
\end{align*}
}

\subsection{Interaction Generation Prompt}

\label{sec:model-inter-full-prompt}

Since this is a multi-choice question answering task, the full prompt for \textbf{models} is as follows:
{\ttfamily
\begin{align*}
\small
& \text{You are mimicking a human.} \\
& \text{You are trying to choose the correct answer to the given question.} \\
& \text{Please ask an assistant sub-questions for help approaching answers.} \\
& \text{In each turn, please only ask one sub-question to interact with an } \\
& \text{assistant. In the sub-questions, please include all necessary} \\
& \text{information, such as the question and options, in the original} \\
& \text{question. If you know the answer, please output "So, the answer} \\
& \text{is: A, B, C, or D."} \\
& \\
& \text{\{QA Question and choices\}} \\
& \text{\{User Model's query: [question 1]\}} \\
& \text{\{Assistant's answer: [answer 1]\}} \\
& \text{\{User Model's query: [question 2]\}} \\
& \text{\{Assistant's answer: [answer 2]\}} \\
& \cdots \\
& \text{\{User Model's query: [question $n$]\}} \\
& \text{\{Assistant's answer: [answer $n$]\}} \\
& \text{\{User Model's final answer\}}
\end{align*}
}

If the current turn reaches the maximum number we set, the system prompt before ``\{Question\}'' looks as follows:
{\ttfamily
\begin{align*}
\small
& \text{Please choose the correct answer to the given question. Please } \\
& \text{output "So, the answer is: A, B, C, or D."}
\end{align*}
}

\subsubsection{QA Question and choices format}

The question prompt for multi-choice questions in MMLU is as follows:
{\ttfamily
\small
\begin{align*}
& \text{<question>} \\
& \text{A. <option A>} \\
& \text{B. <option B>} \\
& \text{C. <option C>} \\
& \text{D. <option D>}
\end{align*}
}
For HotpotQA and AmbigQA datasets, the question prompt only contains a question, such as:
{\ttfamily
\small
<question>
}

\subsection{Persona Prompts}

\label{sec:persona-prompt}

We design distinct prompts for each persona.
Both prompts in meta-evaluation and model interaction modules change with personas.

In model interaction prompts, we only modify the first sentence based on personas. 
See all persona prompts in Table \ref{tab:persona_descriptions}.

\begin{table}[H]
\centering
\caption{Persona Interaction and Evaluation Descriptions}
\begin{tabular}{cp{5.5cm}p{5.5cm}}
\hline

\textbf{Persona} & \textbf{Persona Interaction Description} & \textbf{Persona Evaluation Description} \\ \hline

Expert   &  \ttfamily  You are mimicking a knowledgeable human who can quickly understand new concepts. You need help from an assistant to learn and answer questions. &\ttfamily The AI Assistant helps a knowledgeable human to answer a question. The assistant should provide straightforward, informative, and in-depth answers to human questions. \\ \hline
Critical-Thinker   &  \ttfamily  You are mimicking a human who prefers interactions rich in critical information. You need help from an assistant and try to get critical information from it to answer the following questions. &\ttfamily The AI Assistant should provide clear, non-vague, and precise information or options and help user deduce answers. (Detailed evaluation criteria were indicated but not fully transcribed due to length.) \\ \hline
Adaptability-Seeker &  \ttfamily  You are mimicking a human who prefers an adaptable assistant who can always understand his questions. You need help from an assistant to answer questions. &\ttfamily The AI Assistant helps a human who prefers an adaptable assistant. The assistant should understand user's questions, provide related options, and help user deduce answers. \\ \hline
Clarity-Seeker   &  \ttfamily  You are mimicking a human who prefers clear information in conversations. You need help from an assistant and want to get clear information from it to answer questions. &\ttfamily The AI Assistant helps a human who prefers clear information in conversations. The AI should provide non-vague, precise information to help user deduce answers. \\ \hline
\end{tabular}
\vspace{2mm}
\label{tab:persona_descriptions}
\end{table}

\section{Additional Experiments on Natural Questions}
\label{sec:benchmarking-nq}

We benchmark IQA models in another dataset called Natural Questions (\cite{kwiatkowski-etal-2019-natural}). This dataset comprises authentic questions posed by users about Wikipedia articles, demanding true multi-turn dialogues for resolution, akin to the setup in multi-turn conversational QA dataset QuAC (\cite{choi2018quac}). The experiment results are as in Tables \ref{tab:nq-claude} and \ref{tab:nq-gpt4}. All numbers of queries in the two tables are around 3, and each response to a query from IQA models contains an average of 2 sentences.

Given the number of sentences in each IQA model’s response in Table \ref{tab:nq-non-inter}, the non-interactive outputs are roughly equivalent to about two interaction turns, less than three turns in interactive outputs.
Thus, The interaction process of \ours\ involves not only reasoning processes but also simulating genuine interactive multi-turn conversations. This suggests that the performances shown in tables \ref{tab:nq-claude} and \ref{tab:nq-gpt4} above are driven more by multi-turn interactions than by reasoning processes. Furthermore, these interactions lead to enhanced accuracy, as demonstrated by the superior results in the first two tables compared to those in the last table (non-interactive).

\begin{table}[]
\centering
\caption{IQA-Eval benchmarking results on the Natural Questions by Claude-3}
\resizebox{0.7\columnwidth}{!}{%
\begin{tabular}{l|cccc}
\toprule
IQA models & Helpfulness   & Fluency       & \# Queries     & Accuracy \\
\midrule
GPT3.5     & \textbf{4.86} & \textbf{4.88} & \textbf{2.82} & 0.42     \\
Claude     & 4.88          & 4.90          & 3.02          & 0.38     \\
Llama2     & 4.90          & 4.84          & 3.18          & 0.34     \\
Zephyr     & 4.84          & 4.90          & 3.02          & 0.28     \\
\bottomrule
\end{tabular}%
}
\label{tab:nq-claude}

\caption{IQA-Eval benchmarking results on the Natural Question by GPT-4}
\resizebox{0.7\columnwidth}{!}{%
\begin{tabular}{l|cccc}
\toprule
IQA models & Helpfulness   & Fluency       & \# Queries     & Accuracy \\
\midrule
GPT3.5     & \textbf{4.12} & \textbf{5.00} & \textbf{2.76} & 0.44     \\
Claude     & 4.02          & 5.00          & 2.76          & 0.40     \\
Llama2     & 3.20          & 4.84          & 3.08          & 0.32     \\
Zephyr     & 3.30          & 4.86          & 2.92          & 0.36     \\
\bottomrule
\end{tabular}%
}
\label{tab:nq-gpt4}

\caption{Average number of sentences and accuracy scores of IQA Models (non-interactive setting)}
\resizebox{0.5\columnwidth}{!}{%
\begin{tabular}{l|cc}
\toprule
IQA models & \# Sentences & Accuracy \\
\midrule
GPT3.5     & 4.66        & 0.38     \\
Claude     & 3.16        & 0.34     \\
Llama2     & 5.21        & 0.30     \\
Zephyr     & 4.68        & 0.24     \\
\bottomrule
\end{tabular}%
}
\label{tab:nq-non-inter}
\end{table}

\section{Sensitive to Persona Distributions}
\label{sec:persona-distribution}

We conduct two new experiments to study the effects of changing persona assignments.
Results indicate that \ours\ is sensitive to incorrect persona assignments.
When the persona distribution is incorrect (such as 20\% Expert in Table \ref{tab:persona-distributions}), the performance of \ours\ shows a lower correlation with human evaluations.

Moreover, the last two lines in Table \ref{tab:persona-distributions} describe the correlation between human evaluations and \ours\ within a sub-group only containing pure experts. The correlation results in line “\ours\ (Pure Expert)” represent that (1) our personas accurately represent the pure expert group, as its correlation with the line “Human (Pure Expert)” remains nearly consistent with those in line “\ours\ (Expert)” and (2) given this completely correct persona distribution, our \ours\ correlates well with human evaluations.

\begin{table}[]
\caption{IQA-EVAL results under different persona distribution on the expert persona.}
\resizebox{\columnwidth}{!}{%
\begin{tabular}{l|cccc|cccc}
\toprule
                       & \multicolumn{4}{c|}{Helpfulness}               & \multicolumn{4}{c}{Fluency}                   \\
LEA models             & TDA  & TB   & DA   & $\rho$ & TDA  & TB   & DA   & $\rho$ \\
\midrule
Human                  & 4.60 & 3.84 & 3.52 &                          & 4.35 & 3.84 & 3.22 &                          \\
IQA-EVAL (Expert)      & 4.17 & 3.08 & 3.12 & 0.756                    & 4.47 & 3.84 & 3.40 & 0.787                    \\
IQA-EVAL (20\% Expert)  & 4.31 & 3.26 & 3.44 & 0.708                    & 4.62 & 4.09 & 3.65 & 0.741                    \\
IQA-EVAL (40\% Expert)  & 4.21 & 3.14 & 3.23 & 0.751                    & 4.49 & 3.88 & 3.44 & 0.779                    \\
IQA-EVAL (60\% Expert)  & 4.11 & 3.01 & 3.00 & 0.725                    & 4.43 & 3.77 & 3.34 & 0.734                    \\
IQA-EVAL (80\% Expert)  & 4.02 & 2.90 & 2.79 & 0.680                    & 4.30 & 3.56 & 3.12 & 0.703                    \\
\midrule
Human (Pure Expert)    & 4.69 & 4.00 & 3.73 &                          & 4.36 & 3.96 & 3.26 &                          \\
IQA-EVAL (Pure Expert) & 4.37 & 3.57 & 3.33 & 0.778                    & 4.20 & 3.40 & 2.97 & 0.786                    \\
\bottomrule
\end{tabular}%
}
\label{tab:persona-distributions}
\end{table}

\section{Accurate QA models are preferred by humans in IQA-EVAL}

\label{sec:accurate}

The quote from our cited paper \cite{lee2023evaluating} is ``[...] perception of helpfulness is not necessarily reflected in the overall interaction accuracy.'' It describes the conclusion of multiple tasks in that paper (e.g. text summarization, social dialogue, QA). However, in the QA settings, Table 3 in \cite{lee2023evaluating} shows that humans prefer accurate models on the QA task.

We also conducted experiments (1) using LEA to evaluate interactions between LEAs and IQA models (interactive) and (2) using LEA to evaluate direct answers generated by IQA models (non-interactive). Our experiments in Tables \ref{tab:eval-inter} and \ref{tab:eval-non-inter} show that LEA models prefer accurate models, which aligns well with the conclusion from human annotations.

\begin{table}[]
\caption{Evaluation results of interactions between LEA and IQA models.}
\resizebox{\columnwidth}{!}{%
\begin{tabular}{l|cccc|cccc|cccc}
\toprule
                & \multicolumn{4}{c|}{Helpfulness}                                                                                                           & \multicolumn{4}{c|}{Fluency}                                                                                                               & \multicolumn{4}{c}{Accuracy}                                                                                                              \\
LEA models      & \begin{tabular}[c]{@{}c@{}}GPT\\ 3.5\end{tabular} & \begin{tabular}[c]{@{}c@{}}Claude\\ -instant\end{tabular} & \begin{tabular}[c]{@{}c@{}}Llama2\\ -8b\end{tabular} & \begin{tabular}[c]{@{}c@{}}Zephyr\\ -Alpha\end{tabular} & \begin{tabular}[c]{@{}c@{}}GPT\\ 3.5\end{tabular} & \begin{tabular}[c]{@{}c@{}}Claude\\ -instant\end{tabular} & \begin{tabular}[c]{@{}c@{}}Llama2\\ -8b\end{tabular} & \begin{tabular}[c]{@{}c@{}}Zephyr\\ -Alpha\end{tabular} & \begin{tabular}[c]{@{}c@{}}GPT\\ 3.5\end{tabular} & \begin{tabular}[c]{@{}c@{}}Claude\\ -instant\end{tabular} & \begin{tabular}[c]{@{}c@{}}Llama2\\ -8b\end{tabular} & \begin{tabular}[c]{@{}c@{}}Zephyr\\ -Alpha\end{tabular} \\
\midrule
IQA-EVAL-GPT4   & 4.60    & 4.60                                                      & 3.83      & 4.27                                                    & 4.97    & 5.00                                                      & 4.87      & 4.93                                                    & 0.93    & 0.93                                                      & 0.83      & 0.93                                                    \\
IQA-EVAL-Claude & 4.90    & 5.00                                                      & 4.97      & 4.97                                                    & 4.87    & 5.00                                                      & 4.93      & 4.87                                                    & 0.73    & 0.8                                                       & 0.57      & 0.73                                                    \\
\bottomrule
\end{tabular}%
}
\label{tab:eval-inter}

\caption{Evaluation results of non-interactions (direct answers) between LEA and IQA models.}
\resizebox{\columnwidth}{!}{%
\begin{tabular}{l|cccc|cccc|cccc}
\toprule
                & \multicolumn{4}{c|}{Helpfulness}                                                                                                           & \multicolumn{4}{c|}{Fluency}                                                                                                               & \multicolumn{4}{c}{Accuracy}                                                                                                              \\
LEA models      & \begin{tabular}[c]{@{}c@{}}GPT\\ 3.5\end{tabular} & \begin{tabular}[c]{@{}c@{}}Claude\\ -instant\end{tabular} & \begin{tabular}[c]{@{}c@{}}Llama2\\ -8b\end{tabular} & \begin{tabular}[c]{@{}c@{}}Zephyr\\ -Alpha\end{tabular} & \begin{tabular}[c]{@{}c@{}}GPT\\ 3.5\end{tabular} & \begin{tabular}[c]{@{}c@{}}Claude\\ -instant\end{tabular} & \begin{tabular}[c]{@{}c@{}}Llama2\\ -8b\end{tabular} & \begin{tabular}[c]{@{}c@{}}Zephyr\\ -Alpha\end{tabular} & \begin{tabular}[c]{@{}c@{}}GPT\\ 3.5\end{tabular} & \begin{tabular}[c]{@{}c@{}}Claude\\ -instant\end{tabular} & \begin{tabular}[c]{@{}c@{}}Llama2\\ -8b\end{tabular} & \begin{tabular}[c]{@{}c@{}}Zephyr\\ -Alpha\end{tabular} \\
\midrule
IQA-EVAL-GPT4   & 4.33    & 4.17                                                      & 2.70      & 3.53                                                    & 5.00    & 4.97                                                      & 4.13      & 4.33                                                    & 0.83    & 0.80                                                      & 0.47      & 0.57                                                    \\
IQA-EVAL-Claude & 4.97    & 5.00                                                      & 4.53      & 4.87                                                    & 4.97    & 4.97                                                      & 4.47      & 4.97                                                    & 0.83    & 0.80                                                      & 0.47      & 0.57                                                    \\
\bottomrule
\end{tabular}%
}
\label{tab:eval-non-inter}
\end{table}

\section{Bias Evaluation}
\label{sec:ethics}
We follow the method proposed by \cite{sheng2021revealing} and conduct a new experiment to evaluate the offensiveness and harmfulness of our personas using the RealToxicityPrompts dataset (\cite{gehman2020realtoxicityprompts}) on our LEA models. The results are in the table \ref{tab:ethics}. The values in the table above represent the success rates (higher is better) for each bias metric, persona, and LEA model (\gptthree and \claude). Scores labeled "None" are consistently lower than those for all personas, indicating that our personas do not increase offensiveness or harmfulness in conversations.

\begin{table}[H]
\caption{Evaluating persona biases on offensiveness and harmful metrics. A high score indicates better results.}
\resizebox{\columnwidth}{!}{%
\begin{tabular}{l|cc|cc}
\toprule
                    & \multicolumn{2}{c|}{\textbf{Offensiveness}}         & \multicolumn{2}{c}{\textbf{Harmful}}                \\
\textbf{Persona}    & \textbf{IQA-EVAL-GPT3.5} & \textbf{IQA-EVAL-Claude} & \textbf{IQA-EVAL-GPT3.5} & \textbf{IQA-EVAL-Claude} \\
\midrule
None                & 89.5                     & 91.7                     & 62.5                     & 72.5                     \\
\midrule
Expert              & 95.5                     & 97.3                     & 67.8                     & 75.5                     \\
Critical-Thinker    & 93.3                     & 95.5                     & 65.4                     & 73.7                     \\
Adaptability-Seeker & 94.5                     & 95.5                     & 62.8                     & 74.6                     \\
Clarity-Seeker      & 95.0                     & 94.0                     & 62.5                     & 73.0                     \\
\bottomrule
\end{tabular}%
}
\label{tab:ethics}
\end{table}

\end{document}